\documentclass{nextgame2026}

\usepackage{booktabs}
\usepackage{xcolor}

\usepackage{amsmath,amsfonts,bm}

\def\eqref#1{equation~\ref{#1}}

\def\1{\bm{1}}

\DeclareMathAlphabet{\mathsfit}{\encodingdefault}{\sfdefault}{m}{sl}
\SetMathAlphabet{\mathsfit}{bold}{\encodingdefault}{\sfdefault}{bx}{n}

\def\gA{{\mathcal{A}}}

\def\gO{{\mathcal{O}}}

\def\gR{{\mathcal{R}}}
\def\gS{{\mathcal{S}}}
\def\gT{{\mathcal{T}}}

\def\gZ{{\mathcal{Z}}}

\newcommand{\R}{\mathbb{R}}

\title[EMAgnet]{EMAgnet: Parameter-Space EMA Regularization for Policy Gradient Self-Play in Large Games}

\optauthor{%
\Name{Tristan Maidment\textsuperscript{1}}\Email{}%
\and\Name{JB Lanier\textsuperscript{1}}\Email{}%
\and\Name{Chase McDonald}\Email{}%
\and\Name{Nathan Tsang\textsuperscript{1}}\Email{}%
\and\Name{Eugene Vinitsky\textsuperscript{3}}\Email{},\\
\Name{Roy Fox\textsuperscript{2}}\Email{}%
\and\Name{Albert Wang}\Email{}%
\and\Name{Wesley N. Kerr\textsuperscript{1}}\Email{}%
}

\begin{document}

\maketitle
\vspace{-1em}
\noindent{\small\itshape
\textsuperscript{1}Riot Games \quad \textsuperscript{2}University of California, Irvine \quad \textsuperscript{3}New York University}%
\correspondingnote{Corresponding authors: \texttt{tmaidment@riotgames.com}, \texttt{wkerr@riotgames.com}}
\vspace{0.5em}

\begin{abstract}
Recent work has established that regularized policy gradient methods such as PPO, when used in self-play, can match or exceed specialized game-theoretic algorithms for solving two-player zero-sum imperfect-information games. The uniform distribution has emerged as a strong policy regularization target for this purpose, but it regularizes equally toward all actions regardless of their viability. We introduce EMAgnet, which instead regularizes toward an exponential moving average (EMA) of the last-iterate policy's parameters, providing an adaptive regularization target that evolves with the agent's improving strategy. We evaluate EMAgnet on both standard two-player zero-sum benchmarks and modified benchmarks with exploration challenges and large numbers of strictly dominated strategies. Relative to PPO self-play with uniform-magnet regularization under both linear and power-law annealing schedules, EMAgnet achieves lower exploitability in the majority of tested environments, with consistent performance gains across games containing strictly dominated strategies.
\end{abstract}

\section{Introduction}

Solving two-player zero-sum imperfect-information games (IIGs) has driven notable advances in AI, from Poker \citep{brown2018libratus, brown2020poker} and Stratego \citep{perolat2022mastering, sokota2025stratego} to real-time strategy games like StarCraft \citep{vinyals2019starcraft} and Dota \citep{berner2019dota}. Of these, multiple results have relied on self-play training stabilized through regularization towards target policies \citep{perolat2022mastering, sokota2025stratego}, and recent work has demonstrated that with appropriate regularization, generic policy gradient methods can match or exceed other more specialized game-theoretic approaches \citep{sokota2023mmd, rudolph2025reevaluating}. Given this growing reliance on regularization, the choice of target policy becomes a key design decision.

\begin{figure}[t]
    \centering
    \includegraphics[width=\textwidth]{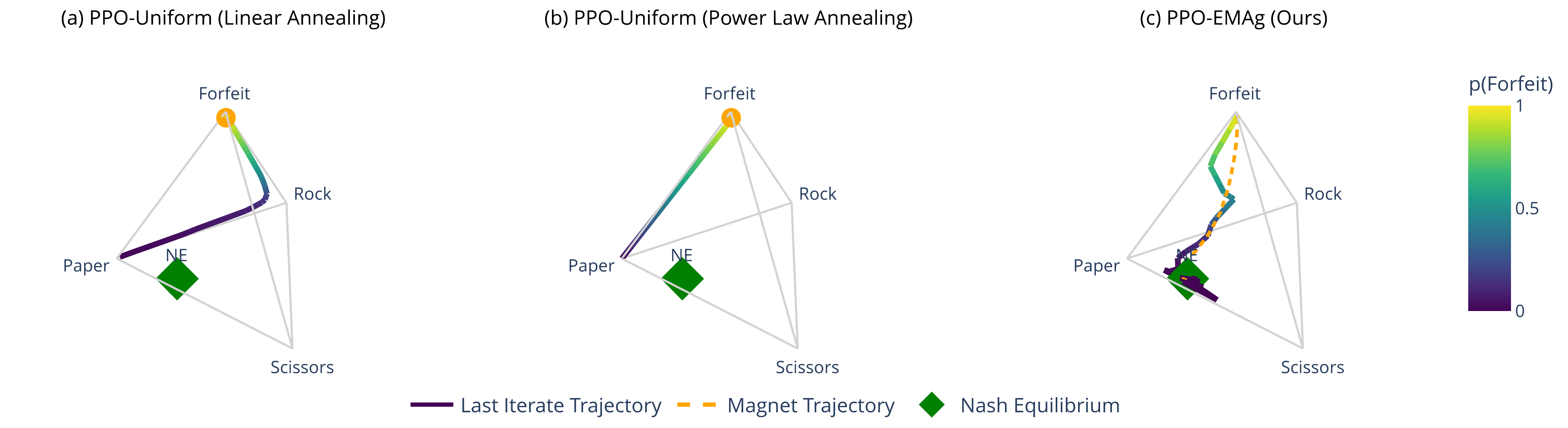}
    \caption{Self-play policy trajectories in Control Biased RPS \citep{lanier2026dags}, where agents must solve gridworld navigation tasks to execute each RPS action or else forfeit. \textbf{(a,b)} Regularizing toward uniform forces the policy to use strictly dominated strategies that fail navigation and forfeit. By the time annealed regularization is weak enough to avoid forfeiting, the policy fails to explore and find the Nash equilibrium (green diamond). \textbf{(c)} PPO-EMAg applies constant regularization toward an EMA of the last-iterate (dashed orange), which regularizes toward viable actions the policy has chosen in the past, enabling convergence.}
    \label{fig:hero}
\end{figure}

\citet{sokota2023mmd} and \citet{rudolph2025reevaluating} establish the uniform distribution, implemented as an entropy bonus, as a straightforward and effective regularization target. \citet{rudolph2025reevaluating} show PPO self-play outperforms more specialized game-solving methods such as R-NaD \citep{perolat2022mastering}, PSRO \citep{lanctot2017psro}, and NFSP \citep{heinrich2016nfsp}. However, the uniform target is strategically agnostic as it regularizes equally toward all strategies regardless of whether they are viable or strictly dominated. In games with large strategy spaces where most options are bad, the uniform target wastes regularization budget on irrelevant strategies (Figure~\ref{fig:hero}a,b). As games grow in complexity, the fraction of the strategy space that is strategically relevant tends to shrink, making this limitation increasingly consequential. In tabular settings, \citet{sokota2023mmd} explored a continuously moving ``magnet'' (the regularization target) that trails behind the current policy, demonstrating faster convergence than annealing uniform regularization strength.

We introduce \textbf{EMAgnet}, which extends this concept to deep RL by maintaining a parameter-space exponential moving average (EMA) of the policy's own network weights as the regularization target (Figure~\ref{fig:hero}c). The EMA magnet continuously adapts to the policy while adding minimal complexity to the standard PPO training loop. As the policy learns to avoid dominated strategies, the magnet also gradually stops regularizing toward them. At the same time, the EMA naturally accumulates a smoothed mixture over strategies encountered during self-play cycling, encouraging the policy to maintain coverage over strategically relevant options. This dual property, forgetting bad strategies while remembering good ones, is what distinguishes EMAgnet from a fixed regularization target.

We extend the tabular moving magnet concept from \citet{sokota2023mmd} to deep RL with PPO~\citep{schulman2017proximalpolicyoptimizationalgorithms} via parameter-space EMA regularization. We denote this new algorithm as \textbf{PPO-EMAg}, and we evaluate it against PPO self-play with uniform-magnet regularization under both linear annealing \citep{rudolph2025reevaluating} and power-law annealing \citep{sokota2025stratego} across standard two-player zero-sum benchmarks and modified benchmarks containing a large number of strictly dominated strategies \citep{lanier2026dags}. PPO-EMAg matches or outperforms uniform baselines on standard benchmarks and outperforms them in games containing strictly dominated strategies.

\section{Preliminaries}

We consider two-player zero-sum games formalized as finite-horizon partially observable stochastic games (POSGs). A game is defined by the tuple
\begin{equation*}\label{eqn:game_tuple}
        \langle \gS, \gA, \gO, \gR, \gT, \Omega, T\rangle,
\end{equation*}
where $\gS$ is the state space, $\gA$ is the action space, $\gO$ is the observation space, $\gR^i:\gS\times\gA\to\R$ is the reward function for player $i$ with $\gR^i = -\gR^{-i}$, $\gT:\gS\times\gA\to\gS$ is the transition function, $\Omega:\gS\to\gO$ is the observation function, and $T$ is the episode horizon. From a sequence of observations and actions, each player constructs an information state $z \in \gZ = \cup_t(\gO\times\gA)^t\times\gO$ that is sufficient for optimal decision-making.

Each player $i \in \{1, 2\}$ acts according to a stochastic policy $\pi_i:\gZ\to\Delta(\gA)$. We measure a joint policy $\pi = (\pi_1, \pi_2)$ by its \textit{exploitability}, the mean incentive across players to deviate to a best response. A joint policy is a Nash equilibrium if and only if its exploitability is zero.

\section{Method}

All methods in this work build on Proximal Policy Optimization \citep{schulman2017proximalpolicyoptimizationalgorithms} in symmetric self-play. We represent the policy with a neural network $\pi_\theta$ and use a shared parameterization for both players, with player identity encoded in the observation. Our baseline, \textbf{PPO-Uniform} \citep{rudolph2025reevaluating}, augments the standard clipped PPO objective $\mathcal{L}_{\text{PPO}}(\theta)$ with an entropy bonus $\lambda_H H(\pi_\theta(\cdot \mid z))$ that regularizes the policy toward the uniform distribution, where $\lambda_H > 0$ may be held fixed or annealed over training. We propose replacing this fixed uniform target with an adaptive one.

\subsection{PPO-EMAg}

The tabular moving magnet of \citet{sokota2023mmd} updates the magnet via a geometric average in policy space at each information state, $\rho_{t+1}(h) \propto \rho_t(h)^{1-\tilde{\eta}} \pi_{t+1}(h)^{\tilde{\eta}}$. With neural network policies, maintaining per-information-state policy averages is impractical. \textbf{PPO-EMAg} instead performs an arithmetic average in parameter space, replacing the uniform magnet with an exponential moving average (EMA) of the policy parameters. The objective becomes
\begin{equation} \label{eqn:emag_loss}
  \mathcal{L}_{\text{PPO-EMAg}}(\theta)
  = \mathcal{L}_{\text{PPO}}(\theta)
  + \lambda_{\mathrm{KL}} \, \mathbb{E}_{z \sim \mathcal{T}}\!\bigl[D_{\mathrm{KL}}\bigl(\pi_{\theta_{\mathrm{mag}}}(\cdot \mid z) \,\|\, \pi_\theta(\cdot \mid z)\bigr)\bigr],
\end{equation}
where $\theta_{\mathrm{mag}}$ denotes the magnet parameters and $\lambda_{\mathrm{KL}} > 0$ controls the regularization strength. After each PPO update, the magnet parameters are updated as
\begin{equation}
  \theta_{\mathrm{mag}} \leftarrow (1 - \tau) \, \theta_{\mathrm{mag}} + \tau \, \theta,
\end{equation}
with step size $\tau \in (0,1]$. The magnet is initialized to the same random weights as the policy, $\theta_{\mathrm{mag}} \leftarrow \theta$. The full training procedure is summarized in Algorithm~\ref{alg:emag} (Appendix).

The key property of PPO-EMAg is that the regularization target adapts with the policy's improving strategy. As the policy learns to avoid strictly dominated strategies, the EMA magnet also gradually stops regularizing toward those strategies. In contrast, the uniform magnet in PPO-Uniform always regularizes toward all actions equally, regardless of their strategic relevance. This distinction becomes significant in games with large strategy spaces containing many suboptimal options, where the uniform magnet wastes regularization budget on irrelevant strategies.

\section{Experiments}

We evaluate PPO-EMAg against two PPO-Uniform baselines across three families of two-player zero-sum games with progressively higher proportions of strictly dominated strategies. Our baselines are PPO-Uniform with linear annealing \citep{rudolph2025reevaluating} and PPO-Uniform with power-law annealing \citep{sokota2025stratego}. All methods share a fixed compute budget and hyperparameter sweep procedure per environment (Appendix~\ref{sec:parameter_sweeps}). For PPO-EMAg, we report exploitability for both the last-iterate policy and the EMA magnet policy, as the magnet often achieves lower exploitability than the last iterate (see Table~\ref{tab:main_results} in the Appendix for full numerical results). We first test on standard games (\S\ref{sec:standard-games}), then on games augmented with strictly dominated strategies (\S\ref{sec:ff-games}), and finally on games where the vast majority of the strategy space is dominated (\S\ref{sec:control-games}).

\begin{figure}[t]
    \centering
    \begin{minipage}[t]{0.27\linewidth}
        \centering
        \includegraphics[width=\linewidth]{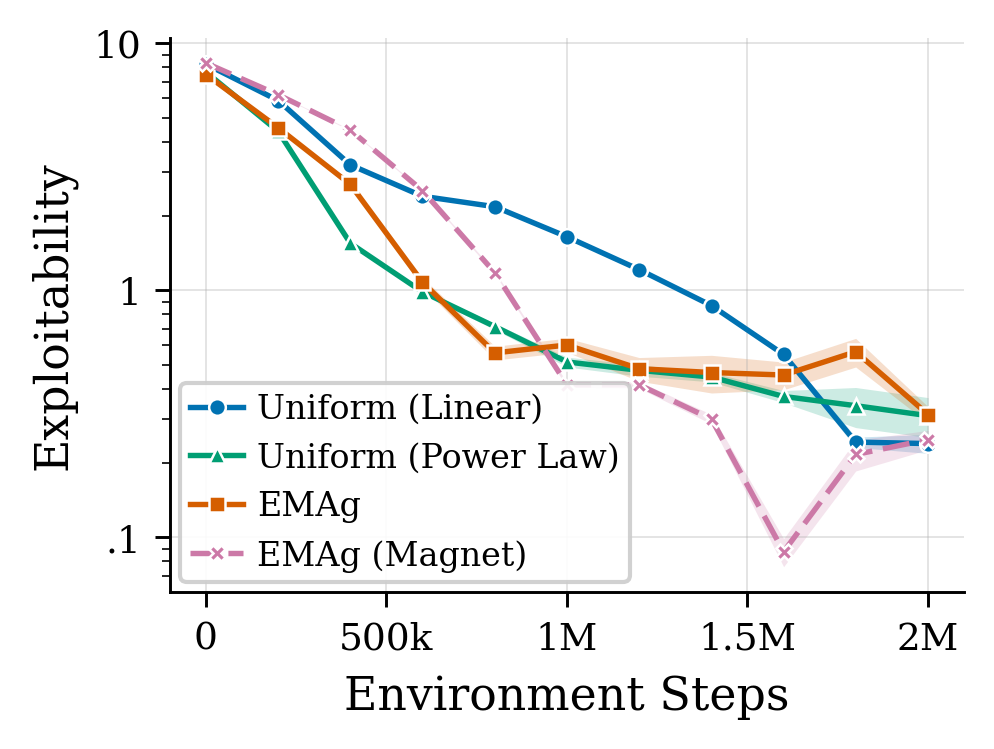}
        \centerline{\scriptsize (a) BRPS}
    \end{minipage}
    \hspace{0.02\linewidth}
    \begin{minipage}[t]{0.27\linewidth}
        \centering
        \includegraphics[width=\linewidth]{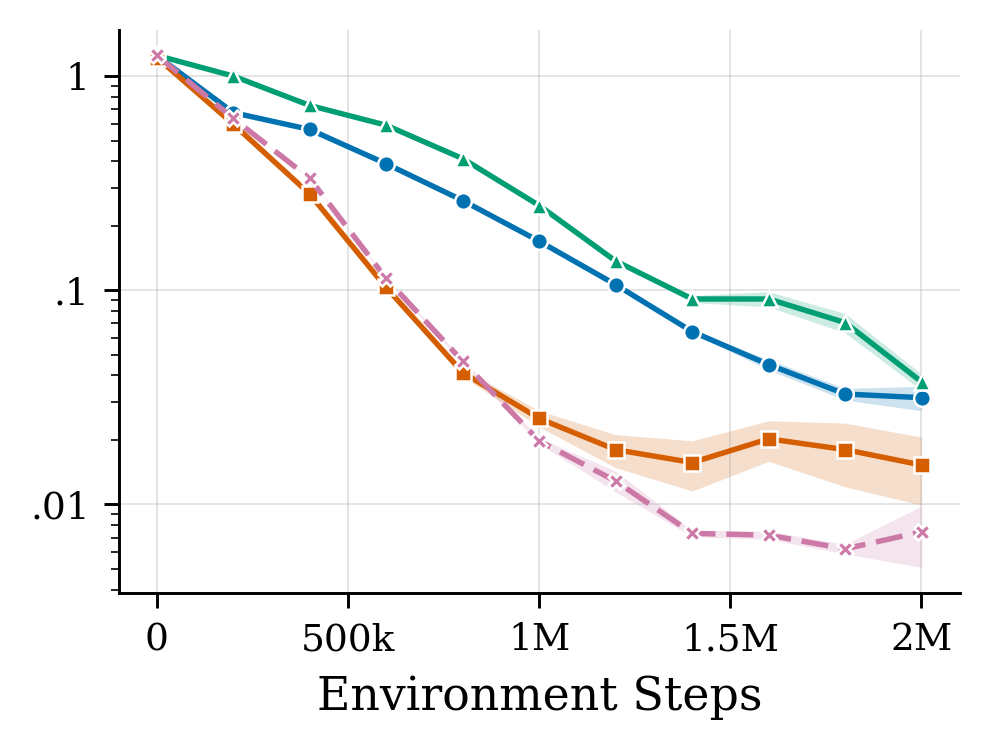}
        \centerline{\scriptsize (b) Goofspiel-4}
    \end{minipage}
    \hspace{0.02\linewidth}
    \begin{minipage}[t]{0.27\linewidth}
        \centering
        \includegraphics[width=\linewidth]{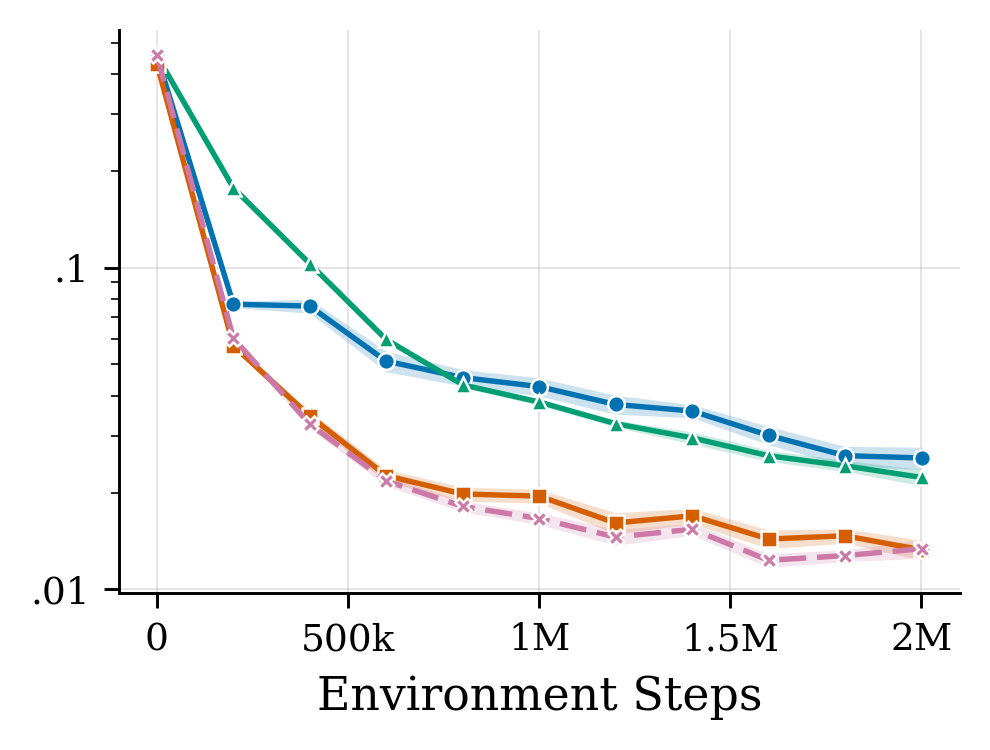}
        \centerline{\scriptsize (c) Kuhn Poker}
    \end{minipage}
    \\[0.2em]
    \begin{minipage}[t]{0.27\linewidth}
        \centering
        \includegraphics[width=\linewidth]{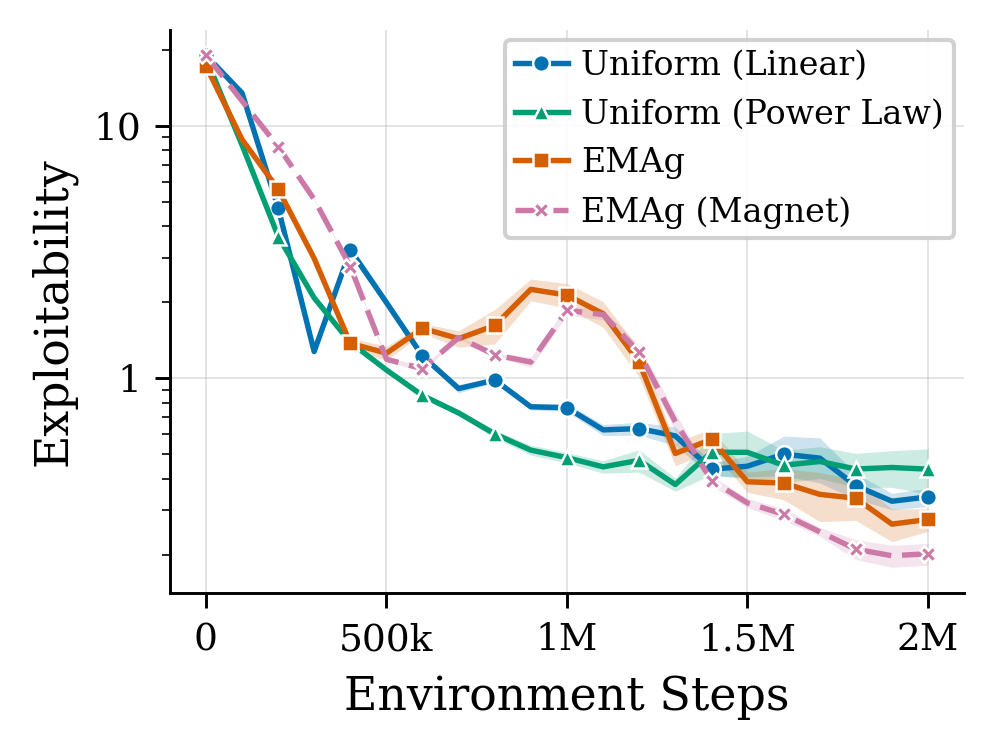}
        \centerline{\scriptsize (d) FF BRPS}
    \end{minipage}
    \hspace{0.02\linewidth}
    \begin{minipage}[t]{0.27\linewidth}
        \centering
        \includegraphics[width=\linewidth]{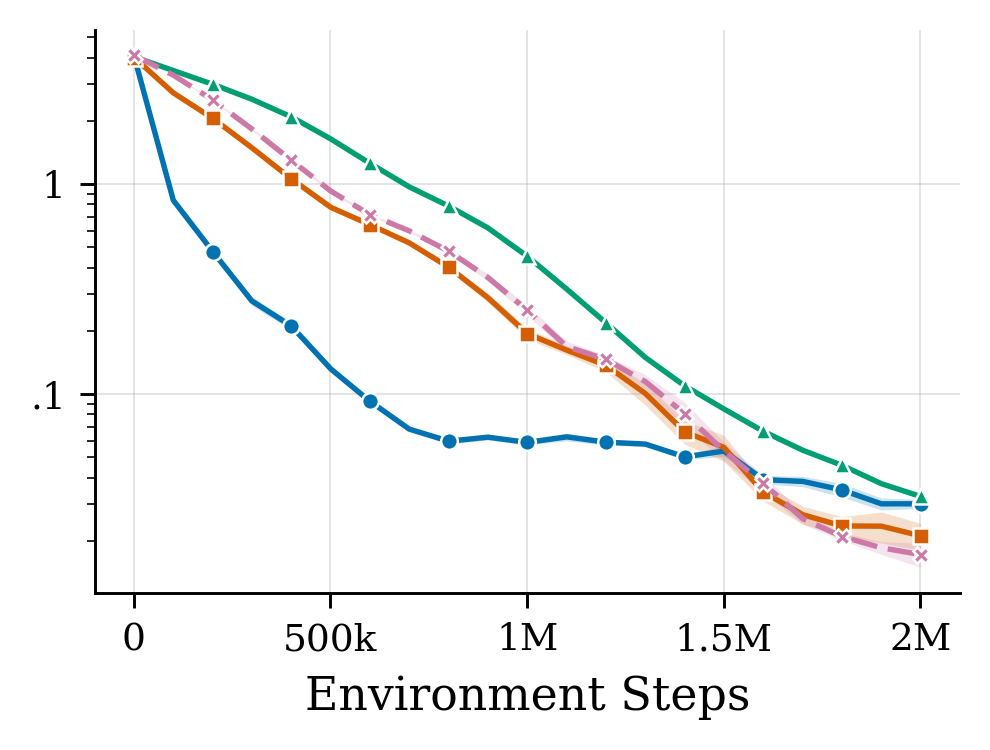}
        \centerline{\scriptsize (e) FF Goofspiel-4}
    \end{minipage}
    \hspace{0.02\linewidth}
    \begin{minipage}[t]{0.27\linewidth}
        \centering
        \includegraphics[width=\linewidth]{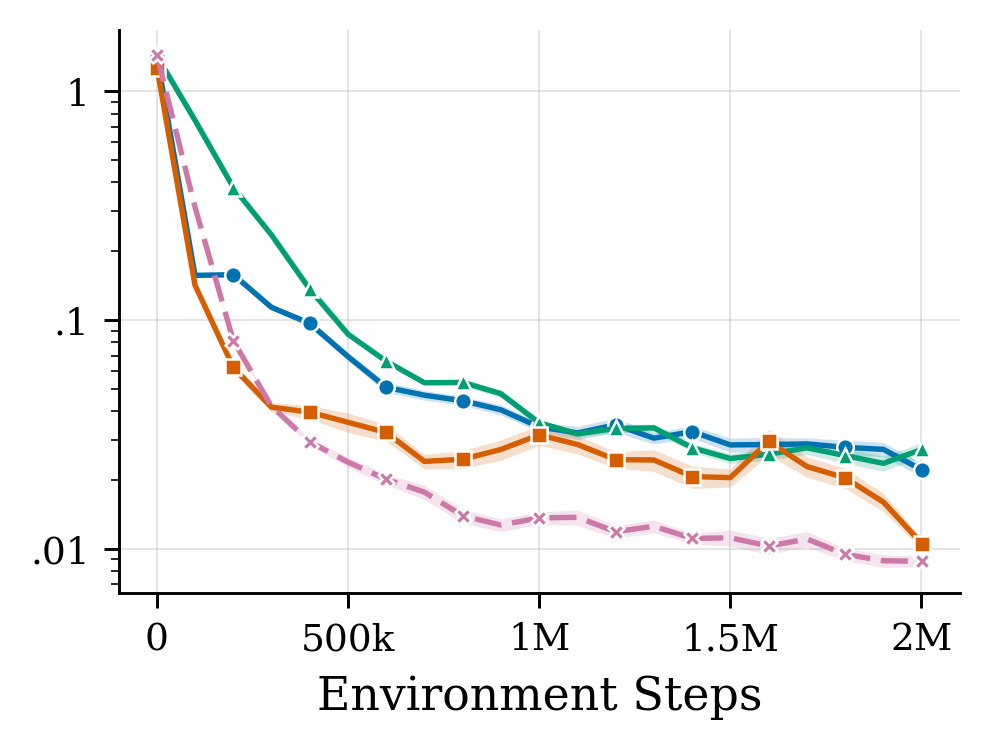}
        \centerline{\scriptsize (f) FF Kuhn Poker}
    \end{minipage}
    \\[0.2em]
    \begin{minipage}[t]{0.27\linewidth}
        \centering
        \includegraphics[width=\linewidth]{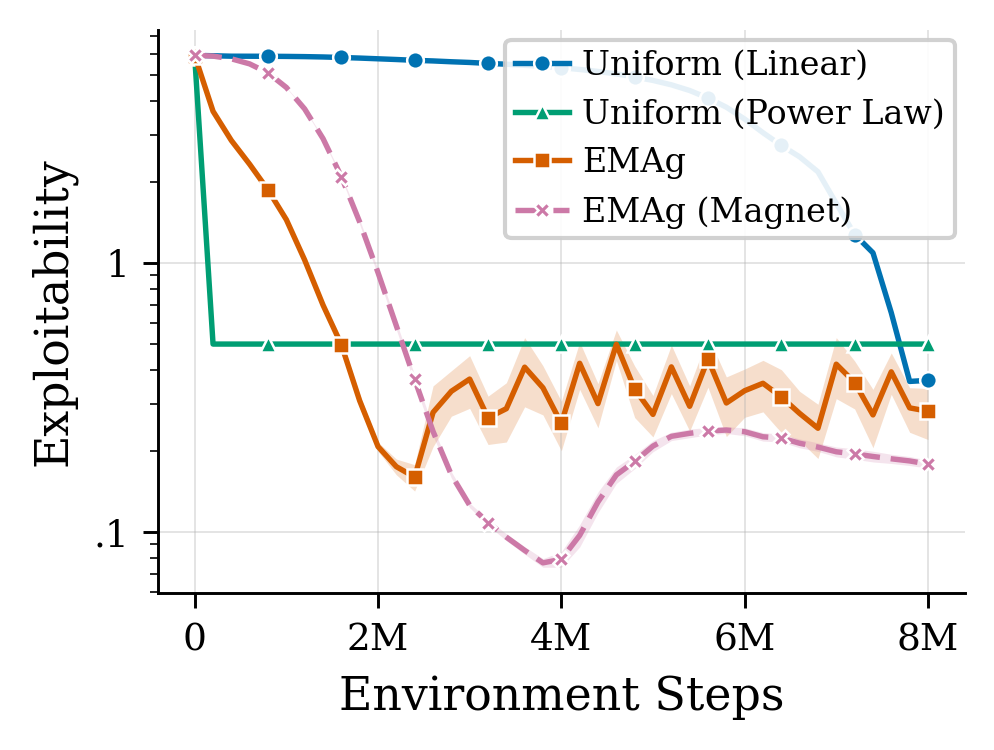}
        \centerline{\scriptsize (g) Control BRPS}
    \end{minipage}
    \hspace{0.02\linewidth}
    \begin{minipage}[t]{0.27\linewidth}
        \centering
        \includegraphics[width=\linewidth]{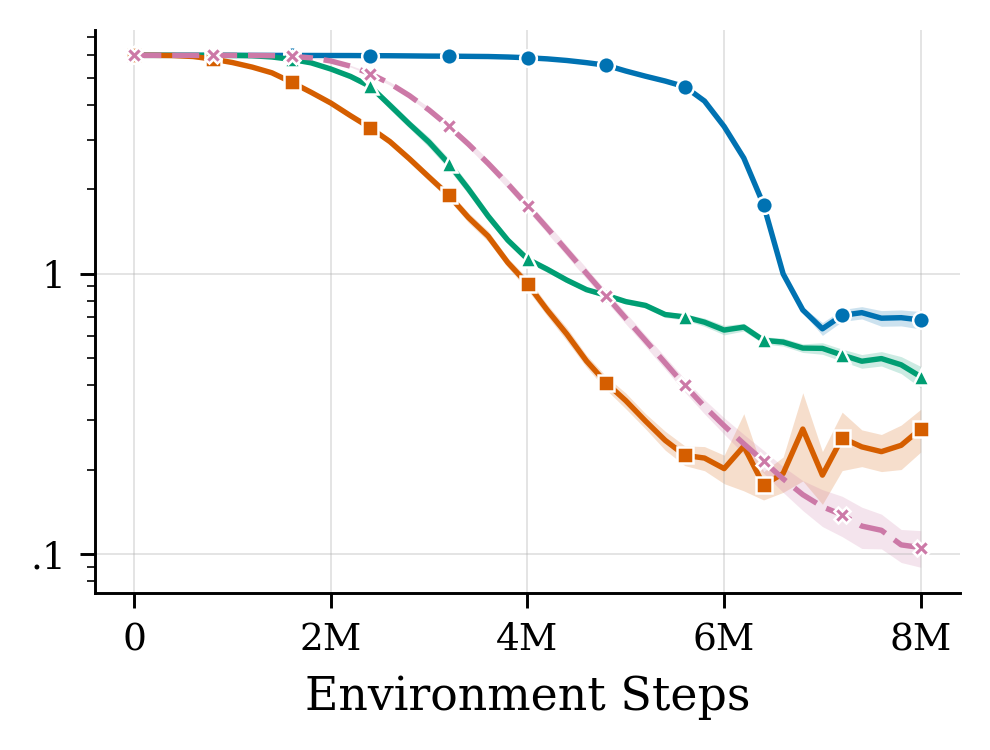}
        \centerline{\scriptsize (h) Control Goofspiel-4}
    \end{minipage}
    \hspace{0.02\linewidth}
    \begin{minipage}[t]{0.27\linewidth}
        \centering
        \includegraphics[width=\linewidth]{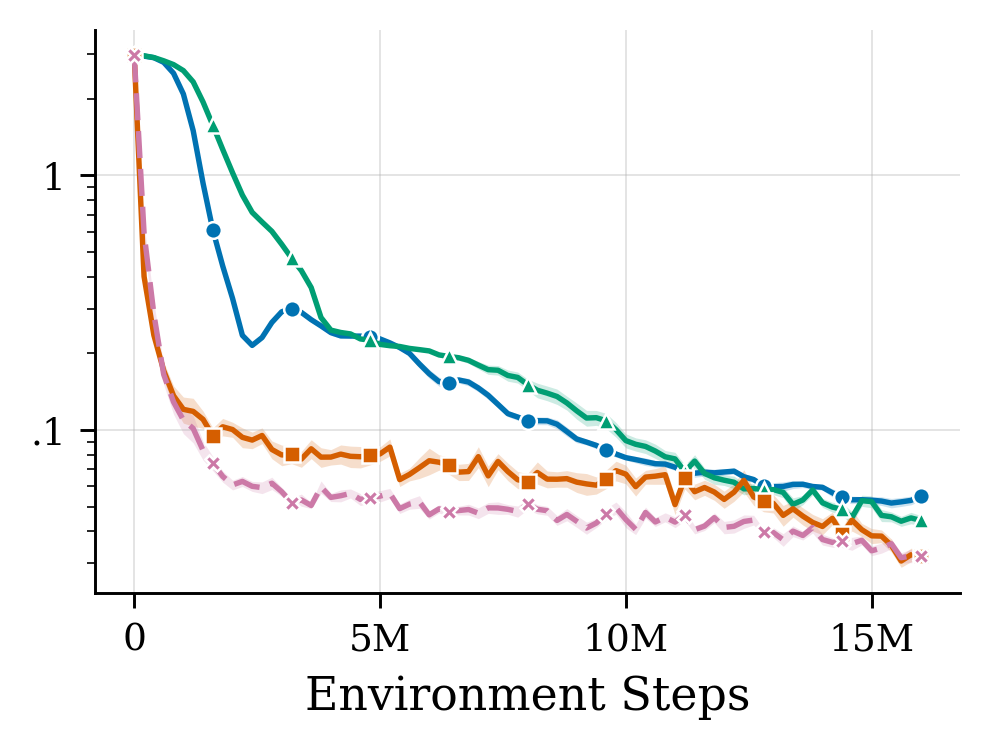}
        \centerline{\scriptsize (i) Control Kuhn Poker}
    \end{minipage}
    \caption{Exploitability over environment steps for each game variant. Top row (a--c): standard games. Middle row (d--f): FF variants with a strictly dominated forfeit action added. Bottom row (g--i): control variants where most strategies are dominated. Best hyperparameter configuration per method (selected via Bayesian sweep), mean across 24 seeds with standard error bands. PPO-EMAg's last-iterate and magnet policies both outperform baselines in FF and control variants, with the magnet consistently reaching lower exploitability than the last iterate. In the control games (g--i), PPO-EMAg also converges significantly faster than both baselines.}
    \label{fig:exploitability-all}
\end{figure}

\subsection{Standard Games}
\label{sec:standard-games}

We evaluate on three two-player zero-sum games with exact exploitability computation from OpenSpiel \citep{LanctotEtAl2019OpenSpiel}: Biased RPS, 4-Card Goofspiel, and Kuhn Poker. Figure~\ref{fig:exploitability-all}(a--c) shows exploitability over environment steps. All methods converge to low exploitability across the three games. In Goofspiel-4 and Kuhn Poker, both the PPO-EMAg last iterate and magnet outperform the baselines, with the magnet achieving the lowest exploitability in both games. In Biased RPS, all methods reach comparable final exploitability. PPO-EMAg is competitive with all baselines in the standard game formulations tested.

\subsection{Forfeit (FF) Games}
\label{sec:ff-games}

We apply the forfeit transformation from \citet{lanier2026dags} to each of the three base games. Every decision node is augmented with a forfeit action. In a game with utilities bounded in $[u_{\min}, u_{\max}]$, the forfeiting player receives $u_{\min} - 1$ and the opponent receives $-(u_{\min} - 1)$, making forfeiting strictly worse than any base-game outcome. The FF variants add strictly dominated strategies to each game while preserving the strategic structure for non-forfeit play.

Figure~\ref{fig:exploitability-all}(d--f) shows exploitability curves for the FF variants. The EMA magnet achieves the lowest final exploitability across all three FF games. The key comparison is between Biased RPS and FF Biased RPS. In standard Biased RPS, PPO-EMAg performs comparably to the baselines. In FF Biased RPS, the only change is the addition of a strictly dominated forfeit action, yet PPO-EMAg now outperforms all baselines. In FF Goofspiel and FF Kuhn, both the last iterate and magnet outperform the baselines, with clear separation by the end of training. The uniform magnet continues to regularize equally toward forfeit throughout training, while the EMA magnet adapts to reduce regularization towards it.

\subsection{Control Games}
\label{sec:control-games}

The control game transformation from \citet{lanier2026dags} embeds each base-game decision within a multi-step gridworld navigation task. At each decision point, the acting player is placed at a starting position on a grid containing one designated action square per available base-game action. The player navigates using directional movement actions (left, right, up, down, stay) for a fixed number of steps. The grid position reached at the end of the timer determines the base-game action taken. If the player is not on any action square when time expires, the forfeit action is selected. Each player acts independently during navigation and cannot observe the opponent's grid position.

Unlike the FF variants, where only a single dominated action is added, the control variants produce a strategy space in which the vast majority of strategies are strictly dominated, as most navigation sequences fail to reach any action square. This structure reflects many real-world competitive games where high-level strategic mixing occurs over a small subset of viable options, but executing each option requires a long sequence of coordinated actions.

Because players act independently during navigation, control-game policies are analytically reducible to their corresponding FF game \citep{lanier2026dags}. For each base-game information state, the navigation policy's induced distribution over base-game actions (and forfeit) can be computed by evaluating it over all grid positions and timer values, yielding an equivalent mixed strategy from which exact exploitability is computed. Full details on the control game configurations used in our experiments are provided in Appendix~\ref{sec:control_game_details}.

Figure~\ref{fig:exploitability-all}(g--i) shows exploitability curves for the control variants. PPO-EMAg outperforms all baselines across all three control environments, with the magnet again achieving the lowest exploitability. The advantage is particularly striking in terms of convergence speed. In Control BRPS (g), Uniform Linear fails to learn for approximately 7M steps before dropping late in training, while Uniform Power Law remains at high exploitability throughout. PPO-EMAg converges by 2M steps. In Control Goofspiel (h), Uniform Linear barely improves over 6M steps while PPO-EMAg reaches low exploitability much earlier. Control Kuhn (i) shows a similar pattern, with PPO-EMAg converging faster than uniform-magnet methods.

Because control-game policies are reducible to the FF game, we can project the learned policies in Control BRPS onto the 4-simplex over \{rock, paper, scissors, forfeit\} and visualize their trajectories (Figure~\ref{fig:hero}). All trajectories start near the forfeit vertex (yellow, high p(forfeit)). With uniform annealing (Figure~\ref{fig:hero}a,b), strong early regularization pulls the policy toward all actions including forfeit, causing it to spend training time on strictly dominated strategies. As regularization anneals toward zero, the policy escapes forfeit but now lacks the regularization pressure needed to explore the full strategy space. It settles near a pure strategy without exploring the rest of the strategy space. This reveals a fundamental dilemma with annealed uniform regularization; when strong, it wastes budget on dominated strategies, and when weak, it provides insufficient force to encourage mixing.

PPO-EMAg (Figure~\ref{fig:hero}c) avoids this dilemma. The policy gradually reduces its mass on the forfeit action with small cycles, never committing entirely to one strategy. Because the EMA magnet (dashed orange) tracks the policy, it stops regularizing toward forfeit once the policy learns to avoid it, while constant (non-annealed) KL regularization continues to encourage mixing throughout training. Once the policy reaches the subspace of non-dominated strategies, it continues cycling near the Nash equilibrium rather than collapsing to a pure strategy.
\section{Discussion}

We introduced EMAgnet, extending the tabular moving magnet concept from \citet{sokota2023mmd} to deep RL with PPO by regularizing toward a parameter-space exponential moving average of the policy's own weights. PPO-EMAg is competitive with uniform-magnet baselines on standard game-solving benchmarks and outperforms them in games containing strictly dominated strategies. Our simplex analysis in Control BRPS (Figure~\ref{fig:hero}) illustrates a limitation of uniform regularization. In games with strictly dominated strategies, a uniform magnet may either waste budget on dominated strategies or lose the regularization force needed for mixing. The EMA magnet avoids this by maintaining its regularization strength, but adapting its target.

As games grow in complexity, the fraction of the strategy space that is strategically relevant tends to shrink. In large-scale competitive games, most possible action sequences are suboptimal, and an agent that allocates regularization budget to these options pays an increasing cost. PPO-EMAg offers a simple mechanism for adapting the regularization target to the agent's evolving strategy, requiring only a single EMA update per training step beyond the standard PPO loop.

Understanding when and why PPO-EMAg is most effective likely depends on factors beyond dominated strategy density. In future work, we plan to investigate how structural properties of games, such as the balance of transitive vs. cyclic structure in a game and the number and nature of cycles, affect the relative benefit of adaptive regularization.

\bibliography{ref}

\appendix
\section{Related Work}

\subsection{Two-Player Zero-Sum Game Solving}

{A central challenge in two-player zero-sum imperfect-information games is that naive self-play with policy gradient methods can cycle or diverge rather than converge to equilibrium \citep{pmlr-v99-cheung19a}. This has motivated a variety of algorithmic frameworks designed to stabilize learning. Population-based methods maintain a growing set of policies and mix over them. Fictitious play \citep{brown1951fp} and its deep RL successor NFSP \citep{heinrich2016nfsp} average over best responses, while PSRO \citep{lanctot2017psro, mcaleer2020pipeline, mcaleer2021xdo, mcaleer2024sppsro} and NeuPL \citep{liu2022neupl} solve an empirical metagame over a policy population. Regret-minimization approaches such as DREAM \citep{steinberger2020dream} and ESCHER \citep{mcaleerescher} adapt counterfactual regret minimization to function approximation. A third family of regularized policy-gradient methods, including NeuRD \citep{omidshafiei2020neurd}, R-NaD \citep{perolat2022mastering}, and magnetic mirror descent \citep[MMD,][]{sokota2023mmd}, stabilizes last-iterate convergence through explicit regularization terms in the policy objective. Our work builds on this last family, proposing a new form of regularization target that adapts over the course of training.}

{\citet{sokota2023mmd} introduced MMD and showed that policy gradients with strong entropy regularization toward a uniform ``magnet'' policy converge to quantal-response equilibria, achieving performance competitive with CFR in tabular settings. They also explored a \textit{moving magnet} variant in which the magnet trails behind the current policy rather than remaining fixed, demonstrating faster convergence than annealing the regularization temperature, though only in tabular settings. \citet{rudolph2025reevaluating} subsequently demonstrated that generic policy-gradient methods such as PPO, when run in a high uniform regularization regime, are competitive with or superior to all FP-, DO-, and CFR-based deep RL approaches across five large imperfect-information games. This result established PPO self-play with uniform-magnet regularization as a strong and simple baseline. More recently, \citet{sokota2025stratego} achieved superhuman performance in Stratego by, among other innovations, annealing regularization coefficients according to power-law schedules over training, which avoids premature entropy collapse while permitting stronger convergence late in training. R-NaD \citep{perolat2022mastering} takes a different approach, regularizing via reward shaping toward a periodically updated reference policy. At scale, DeepNash gradually transitions between regularization targets using linear interpolation and uses an EMA of the policy parameters to approximate fixed points. However, the regularization targets themselves remain discrete snapshots set at iteration boundaries, and the algorithm requires a complex multi-phase structure with separate dynamics and update stages.}

{Existing regularization targets thus range from fixed and strategically uninformed (the uniform distribution) to adaptive but discrete (R-NaD's periodic snapshots). The tabular moving magnet of \citet{sokota2023mmd} demonstrated that a continuously moving target can yield faster convergence than a fixed uniform one. We propose a parameter-space EMA that provides continuous adaptation with minimal additional complexity over the uniform baseline, extending the moving magnet concept to deep RL.}

\subsection{Weight Averages in Deep Learning}

{Weight-space averaging has had significant empirical successes in deep learning and its use has taken many forms. For example, in supervised learning benchmark tasks~\citet{izmailov2018averaging} demonstrate that their approach (Stochastic Weighted Averaging; SWA) of averaging the weights at specified intervals during stochastic gradient descent resulted in a policy that exhibited improved generalization. \cite{wortsman2022model} built on this approach by demonstrating how an average of diverse fine-tuned policies (``model soups'') improves both performance and generalization.}

{EMAs have become an increasingly common approach to weight averaging. \citet{morales2024exponential} provide a detailed study of the behavior and training dynamics of EMA models. Their work showed that EMA models often exhibit improved generalization and robustness, as well as calibration, consistency, and performance in transfer learning. Beyond the performance of the policies themselves, utilizing the EMA as policy for regularization has been shown to improve performance, stability, and plasticity across domains, such as in reinforcement learning from human feedback~\citep[e.g.,][]{rame2024warp, zhang2026ema} and standard single- and multi-agent reinforcement learning~\citep[e.g.,][]{lillicrap2020continuous, lee2024slow, chen2026enhancing}.}

\section{PPO-EMAg Training Procedure}
\label{sec:algorithm}

EMAgnet requires only a few simple modifications to standard policy gradient methods. We describe the full procedure for PPO-EMAg in~Algorithm~\ref{alg:emag}. At initialization, the magnet policy is initialized with a copy of the randomly initialized parameters $\theta$ from the behavior policy. Data collection remains unchanged from standard PPO, with the only remaining augmentations being the incorporation of the KL loss term in Equation~\ref{eqn:emag_loss}. At the end of each epoch, the parameters of the magnet policy $\theta_{\textrm{mag}}$ are updated via the EMA update using the current $\theta$. 

\begin{algorithm2e}[h]
  \caption{PPO-EMAg self-play}
  \label{alg:emag}
  \SetAlgoLined
  \KwIn{POSG $\mathcal{G}$, joint policy $\pi_\theta$, magnet parameters $\theta_{\mathrm{mag}} \leftarrow \theta$, EMA step size $\tau$, PPO hyperparameters}
  \While{not converged}{
    Initialize empty trajectory buffer $\mathcal{T}$\;
    \For{episode $= 1,\dots,K$}{
      Roll out self-play episode using $\pi_\theta$\;
      Add trajectory data to $\mathcal{T}$\;
    }
    \For{PPO epoch $= 1,\dots,E$}{
      Update $\theta$ using $\mathcal{L}_{\text{PPO-EMAg}}(\theta)$ on minibatches from $\mathcal{T}$\;
      $\theta_{\mathrm{mag}} \leftarrow (1 - \tau)\,\theta_{\mathrm{mag}} + \tau\,\theta$ \tcp*{EMA magnet update}
    }
  }
\end{algorithm2e}

\section{Summary of Results}
\label{sec:results_table}

The final numerical values for exploitability across tasks and algorithms is shown in Table~\ref{tab:main_results}. The results for the last iterate and magnet policy are shown separately, and we compare to uniform regularization baselines. 

There are no statistically significant differences for results in Biased RPS. In all remaining environments, at least one of the two PPO-EMAg policies achieves the lowest exploitability by a statistically significant margin. 

In several cases we observe a significant gain by the magnet over the last iterate policy, although it is not consistent across all settings. In Biased RPS, FF Biased RPS, FF Kuhn Poker, and Control Goofspiel we observe significantly lower exploitability in the magnet relative to the last iterate. The strong performance of the EMA itself is in line with prior empirical work that has demonstrated strong performance in a number of tasks from EMA policies~\citep{morales2024exponential}.

\begin{table}[h!]
\centering
\caption{{Exploitability (lower is better) across standard, FF, and control game benchmarks. Reported as mean $\pm$ 95\% CI (Student's $t$, $n=24$ seeds).  A $\dagger$ on the magnet cell indicates that it differs significantly ($p<0.05$) from the last iterate.}}
\begin{tabular}{lcccc}
\toprule
\textbf{Game} & \textbf{Uniform} & \textbf{Uniform} & \textbf{PPO-EMAg} & \textbf{PPO-EMAg} \\
 & (Linear) & (Power Law) & (Last Iterate) & (Magnet) \\
\midrule
\multicolumn{5}{l}{\textit{Standard Benchmarks}} \\
\hspace{1em}Biased RPS & $\mathbf{0.240 \pm 0.043}$ & $0.311 \pm 0.118$ & $0.314 \pm 0.049$ & $0.249 \pm 0.046^{\dagger}$ \\
\addlinespace[2pt]
\hspace{1em}Goofspiel & $0.031 \pm 0.009$ & $0.037 \pm 0.007$ & $\mathbf{0.015 \pm 0.011}$ & $\mathbf{0.007 \pm 0.005}$ \\
\addlinespace[2pt]
\hspace{1em}Kuhn Poker & $0.026 \pm 0.004$ & $0.022 \pm 0.003$ & $\mathbf{0.013 \pm 0.002}$ & $\mathbf{0.013 \pm 0.002}$ \\
\midrule
\multicolumn{5}{l}{\textit{FF Game}} \\
\hspace{1em}FF Biased RPS & $0.338 \pm 0.058$ & $0.436 \pm 0.184$ & $0.275 \pm 0.058$ & $\mathbf{0.201 \pm 0.041}^{\dagger}$ \\
\addlinespace[2pt]
\hspace{1em}FF Goofspiel & $0.030 \pm 0.003$ & $0.033 \pm 0.002$ & $\mathbf{0.021 \pm 0.006}$ & $\mathbf{0.017 \pm 0.005}$ \\
\addlinespace[2pt]
\hspace{1em}FF Kuhn Poker & $0.022 \pm 0.003$ & $0.027 \pm 0.004$ & $\mathbf{0.011 \pm 0.001}$ & $\mathbf{0.009 \pm 0.001}^{\dagger}$ \\
\midrule
\multicolumn{5}{l}{\textit{Control Games}} \\
\hspace{1em}Control BRPS & $0.367 \pm 0.010$ & $0.500 \pm 0.000$ & $0.281 \pm 0.125$ & $\mathbf{0.179 \pm 0.013}$ \\
\addlinespace[2pt]
\hspace{1em}Control Goofspiel & $0.683 \pm 0.101$ & $0.429 \pm 0.078$ & $\mathbf{0.280 \pm 0.100}$ & $\mathbf{0.105 \pm 0.033}^{\dagger}$ \\
\addlinespace[2pt]
\hspace{1em}Control Kuhn & $0.055 \pm 0.004$ & $0.044 \pm 0.004$ & $\mathbf{0.032 \pm 0.004}$ & $\mathbf{0.032 \pm 0.004}$ \\
\bottomrule
\end{tabular}
\label{tab:main_results}
\end{table}

\section{Hyperparameter Sweeps}~\label{sec:parameter_sweeps}

 To ensure a fair comparison across algorithms and environments, we allocate a standardized and fixed compute budget to hyperparameter searches for each method. This means that our claims and results do not preclude further improvements in any of the algorithms tested, but rather suggest that any further gains in performance or reductions in exploitability require additional resources. 

We use Bayesian optimization for hyperparameter selection, choosing configurations that minimize final exploitability. We sweep over a specified set of parameters for each algorithm, allocating 700 samples from the parameter space with three random seeds per parameter set. All runs are trained to completion (i.e., no early stopping). The parameters tuned for each algorithm are shown in Table~\ref{tab:swept_params}.

For power law schedules, we use the following formulation to scale the entropy and learning rates:
\begin{equation} \label{eqn:power_law}
    \left(1 + \rho C \right)^{-q}, 
\end{equation} 
where $C$ is a scaling coefficient, $q$ is the power law exponent, and $\rho$ is the current training progress in terms of the fraction of total updates completed.

\begin{table}[h]
\centering
\caption{Parameter specifications for each algorithm used across environments.}
\begin{tabular}{lcc}
\toprule
\textbf{Parameter} & Values & \textbf{Sampling Method} \\
\midrule
\multicolumn{3}{l}{\textit{Uniform (Linear)}} \\
\hspace{1em}Entropy Coefficient       & [1e-4, 32.0] & Log Uniform \\
\hspace{1em}Anneal Entropy &  \{true, false\} & Choice \\
\hspace{1em}Learning Rate &  [1e-7, 0.01] & Log Uniform \\
\hspace{1em}Anneal Learning Rate         & \{true, false\} & Choice \\
\hspace{1em}Clip Coefficient        &  [1e-4, 0.4] & Log Uniform \\
\hspace{1em}GAE~$\lambda$ &  [0.6, 1.0] & Uniform \\
\hspace{1em}VF Coefficient &  [0.1, 5.0] & Uniform \\

\midrule
\multicolumn{3}{l}{\textit{Uniform (Power Law)}} \\
\hspace{1em}Learning Rate Power Law $C$       & [1, 1e5] & Log Uniform \\
\hspace{1em}Learning Rate Power Law $q$       & [0.1, 2.0] & Uniform \\
\hspace{1em}Entropy Power Law $C$       & [1, 1e5] & Log Uniform \\
\hspace{1em}Entropy Power Law $q$       & [0.1, 2.0] & Uniform \\
\hspace{1em}Entropy Coefficient       & [1e-4, 32.0] & Log Uniform \\
\hspace{1em}Learning Rate &  [1e-7, 0.01] & Log Uniform \\
\hspace{1em}Clip Coefficient        &  [1e-4, 0.4] & Log Uniform \\
\hspace{1em}GAE~$\lambda$ &  [0.6, 1.0] & Uniform \\
\hspace{1em}VF Coefficient &  [0.1, 5.0] & Uniform \\

\midrule
\multicolumn{3}{l}{\textit{PPO-EMAg}} \\
\hspace{1em}EMAg $\lambda_{\textrm{KL}}$       & [0.01, 32.0] & Log Uniform \\
\hspace{1em}EMAg $\tau$       & [1e-5, 0.1] & Log Uniform \\
\hspace{1em}Entropy Coefficient       & [1e-4, 0.1] & Log Uniform \\
\hspace{1em}Anneal Entropy &  \{true, false\} & Choice \\
\hspace{1em}Learning Rate &  [1e-7, 0.01] & Log Uniform \\
\hspace{1em}Anneal Learning Rate         & \{true, false\} & Choice \\
\hspace{1em}Clip Coefficient        &  [1e-4, 0.4] & Log Uniform \\
\hspace{1em}GAE~$\lambda$ &  [0.6, 1.0] & Uniform \\
\hspace{1em}VF Coefficient &  [0.1, 5.0] & Uniform \\
\bottomrule
\end{tabular}
\label{tab:swept_params}
\end{table}

\section{Control Game Details}
\label{sec:control_game_details}

We apply the forfeit and control game transformations from \citet{lanier2026dags} to Biased RPS, 4-Card Goofspiel, and Kuhn Poker. In each control variant, the base game is first augmented with a forfeit action (the FF transformation), then each decision node is replaced with a gridworld navigation task (the control transformation).

\textbf{Observation space.} At each navigation step, the acting player observes a vector consisting of their normalized grid position (2 dimensions), the fraction of time remaining (1 dimension), and the base-game information state tensor. The opponent's grid position is not observed, which is what enables the analytic reduction to the FF game.

\textbf{Action space.} During navigation, the player selects from five movement actions: left, right, up, down, and stay. The time limit equals the number of steps available to reach an action square. If the player is not on an action square when the timer expires, the forfeit action is taken in the base game.

\textbf{Exploitability computation.} Because players act independently during navigation, the control policy's induced distribution over base-game actions can be computed exactly. For each base-game information state, we evaluate the navigation policy over all grid positions and timer values to compute the probability of terminating on each action square or forfeiting. These probabilities define an equivalent mixed strategy over base-game actions plus forfeit, from which we compute exact exploitability in the base (FF) game. The cost of this reduction is proportional to the number of information states, grid cells, and timer steps, and is orders of magnitude smaller than the cost of self-play training.

\textbf{Configurations used.} Table~\ref{tab:control_configs} lists the specific grid configurations used for each control game in our experiments.

\begin{table}[h]
\centering
\caption{Control game configurations. Each action square corresponds to one base-game action. Forfeit is the default when no action square is reached.}
\begin{tabular}{lcccc}
\toprule
\textbf{Game} & \textbf{Grid Size} & \textbf{Timer (steps)} & \textbf{Dist. to Action Sq.} & \textbf{Action Squares} \\
\midrule
Control BRPS & $5 \times 5$ & 4 & 4 & 3 (rock, paper, scissors) \\
Control Goofspiel-4 & $7 \times 7$ & 5 & 5 & 4 (one per card) \\
Control Kuhn & $5 \times 5$ & 4 & 4 & 2 (bet, pass) \\
\bottomrule
\end{tabular}
\label{tab:control_configs}
\end{table}

\end{document}